\documentclass[11pt]{article}
\usepackage[]{winlp}
\usepackage{times}
\usepackage{latexsym}
\usepackage[normalem]{ulem}
\useunder{\uline}{\ul}{}
\usepackage{subfig}
\usepackage{booktabs}
\usepackage{adjustbox}

\usepackage{microtype}
\usepackage[disable]{todonotes}
\usepackage{enumerate}
\usepackage{enumitem}
\usepackage{tabularx}
\usepackage{comment}

\winlpfinalcopy 

\title{Discovering changes in birthing narratives during COVID-19}

\author{
Daphna Spira$^{1}$  Noreen Mayat$^{1}$  Caitlin Dreisbach$^{2}$   Adam Poliak$^{1,2}$\\
$^{1}$Barnard College\\ 
$^{2}$Columbia Data Science Institute\\
\texttt{\{nm3224,des2188,apoliak\}@barnard.edu c.dreisbach@columbia.edu} 
  }
  
\date{}
\begin{document}
\maketitle

\section{Introduction}
Social media 
has the potential to elucidate a deeper level of understanding of COVID-19's impacts across wide ranging communities. Recent work leveraged publicly available data from social media platforms to detect emerging symptoms from COVID-19~\cite{santosh-etal-2020-detecting} as well as the pandemic's impacts on mental health~\cite{biester-mental-health,tabak-purver-2020-temporal,thukral-etal-2020-identifying,wolohan-2020-estimating}.

With tightened restrictions in hospital settings, COVID-19 has greatly
impacted expecting parents, newborns, 
and their families~\cite{findeklee2020clinical,10.3389/fsoc.2021.611212,10.3389/fsoc.2021.655401,doi:10.1177/23333936211006397,vazquez2021impact,marino2021giving}. 
However, no existing work has used natural language processing techniques to analyze impacts of COVID-19 from long-form text written by new parents explicitly describing their birthing experiences.

We investigate whether, and if so how, birthing narratives written by new parents on Reddit changed during COVID-19. Our results indicate that the presence of family members significantly decreased 
and themes related to induced labor significantly increased in the narratives during COVID-19. 
Our work builds upon recent research that analyze how new parents use Reddit to describe their 
birthing experiences~\cite{fathers-reddit,antoniak2019narrative,mothers-reddit}.

\section{Data Collection}
Reddit is a social media platform where users can post anonymous submissions and comments in various subreddits. We use the Pushshift Reddit API~\cite{Baumgartner2020ThePR} to collect all submissions posted to nine subreddits related to the birthing experience between 
April 2009 and
June 2021.\footnote{These nine subreddits are: 
r/BabyBumps, r/beyondthebump, r/BirthStories, r/daddit, r/predaddit, r/pregnant, r/Mommit, r/NewParents, and r/InfertilityBabies.}
Following \newcite{antoniak2019narrative}, we remove all posts that do not include any of the terms ``birth story,'' ``birth stories,'' or ``graduat,'' guaranteeing our corpus consists of birthing narratives. 
We remove all posts that contain less than $500$ tokens,\footnote{We noticed that in
recent years, new parents post a picture of the baby and then describe the birthing narrative in the first comment. We include these examples in our filtering process.} 
resulting in a corpus of 
4,484 birthing narratives before and 913 during COVID-19.\footnote{We demarcate March 11th, 2020 as the start of 
COVID-19, the day the WHO declared a global pandemic.} 

\section{Method}

\paragraph{Topic Modeling.}
To discover distinct topics across our collection of birthing narratives, we apply Latent Dirichlet Allocation~\cite{blei2003latent}, as implemented in Mallet~\cite{McCallumMALLET}. 
In initial experiments, $k$, the number of topics, ranges from $5$ to $50$. We choose $k=50$ based on $C_{v}$ coherence~\cite{10.1145/2684822.2685324}. 
Discovered topics include induced labor, family, breastfeeding, and the first moments between a new parent and child. 
To determine whether the prevalence of these topics changed during COVID-19, we fit topic-specific Prophet models, an additive regression approach for forecasting time series data~\cite{taylor2018forecasting},
on the topic's average monthly prevalence in our corpus prior to March 2020. We then compare the topic's actual average monthly prevalence in our corpus during COVID-19 with the corresponding model's forecast. Following \newcite{biester-mental-health}, we quantify how often the actual topic's monthly probabilities fall outside the model's 95\% CI and use one-tailed Z-tests to determine statistical significance.

\paragraph{Quantifying Personas Presence.} 
 
Determining the prevalence of types of characters, or personas, in a narrative can illuminate information from an author's perspective, e.g. 
who is most the relevant, valued, or supportive character. 
Following \newcite{antoniak2019narrative}, we quantify a persona's prevalence by counting how often they are mentioned, using a dictionary of terms for each persona.\footnote{
Doctor: [doctor, dr, doc, ob, obgyn, gynecologist, physician]; Partner: [partner, husband, wife]; Nurse: [nurse]; Midwife: [midwife]; Family: [mom, dad, mother, father, brother, sister]; Anesthesiologist: [anesthesiologist]; Doula: [doula]}
 We examine the difference in average 
 mentions of each persona before and during COVID-19.\footnote{Since narratives during COVID-19 were on average roughly 20\% shortner, we adjust the counts. We apply t-tests to compute statistical significance.}

\section{Results}

\begin{figure*}[t!]%
    \centering
    \subfloat[\centering The family members topic fell significantly below  the confidence interval in the post-COVID-19 period.]{\includegraphics[width=1\textwidth]{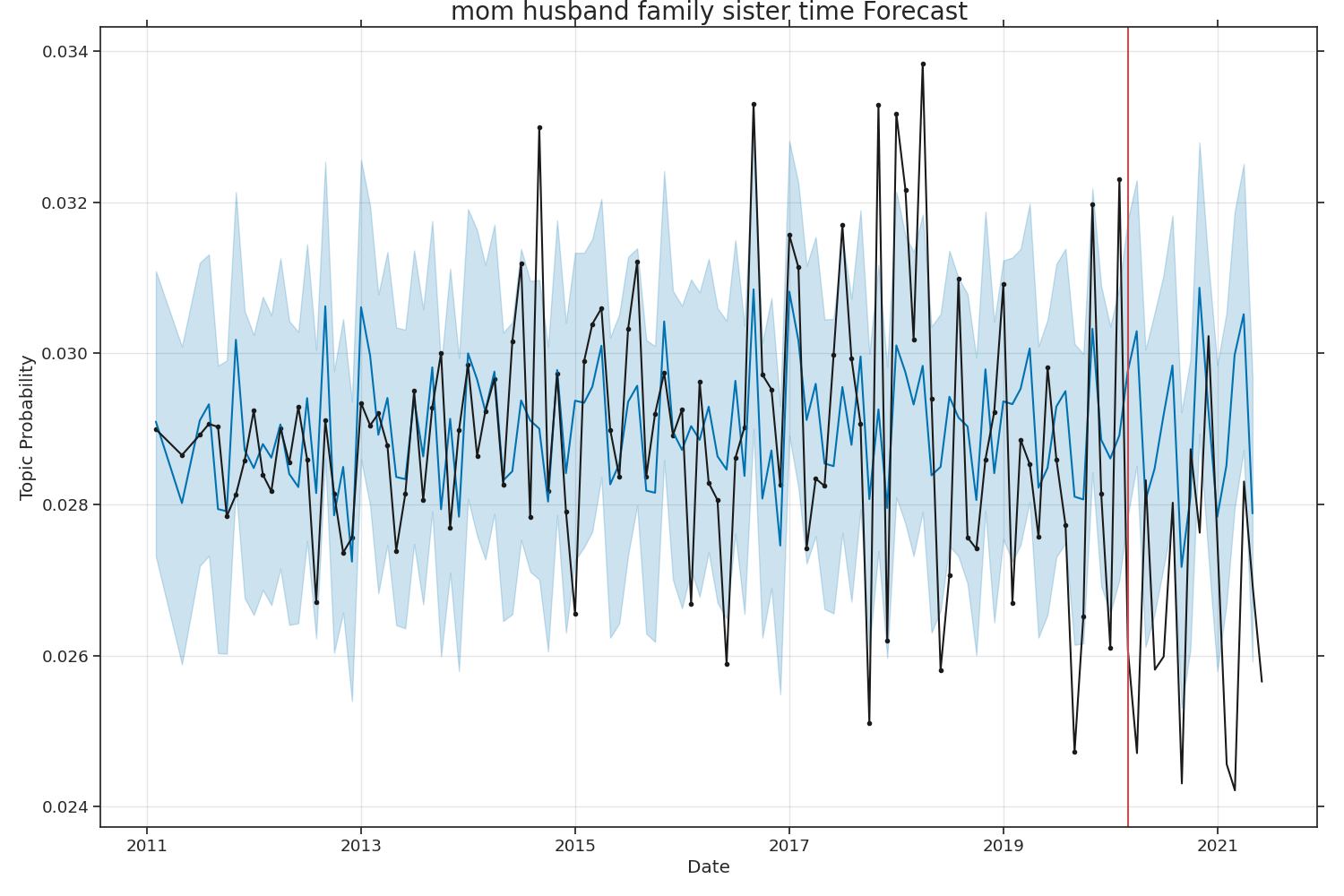}\label{topics:family}}%
    \qquad
    \subfloat[\centering The induction topic rose significantly above the confidence interval in the post-COVID-19 period.]{{\includegraphics[width=1\textwidth]{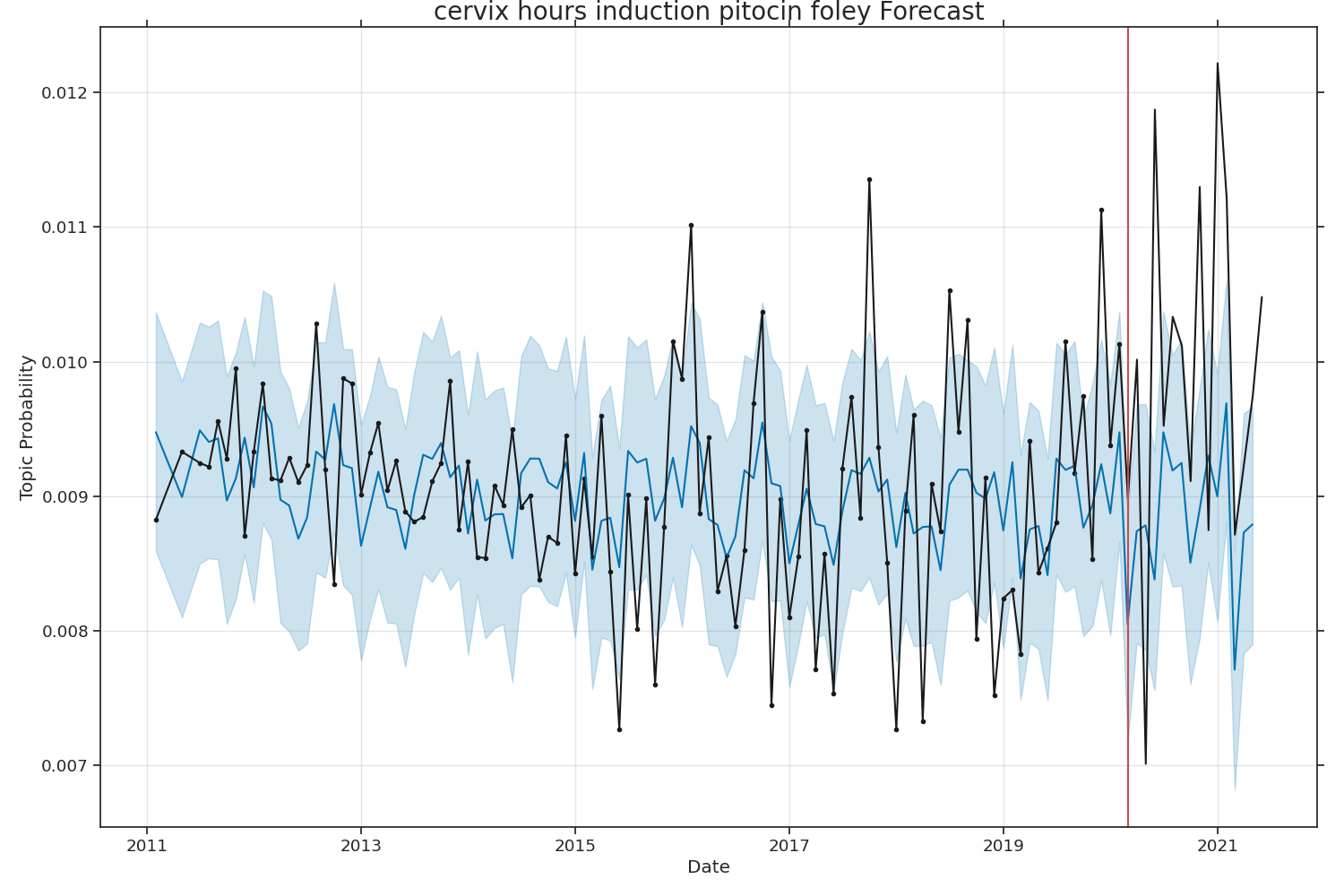}}\label{topics:induction}}%
    \caption{Topic forecasts versus actual data for two topics over time. The x-axis, y-axis, and red vertical line respectively indicate the date, monthly average topic probability, and beginning of COVID-19. 
    The blue line, shaded region, and black line respectively represent the models' prediction, 95\% CI, and actual data. Each figure includes the top-5 words for each topic.} 
    \label{fig:topics}%
\end{figure*}

Figure \ref{fig:topics} shows how the forecasted prevalence's for the \textit{family} and \textit{induction}  topics significantly differ with the topics' prevalence during COVID-19. The increase in the induction topic (Figure \ref{topics:induction}) may reflect the increased recommendation of planned induction, enabling COVID-19 testing of expecting parents in advance of delivery~\cite{goercovid}. The decrease in the family topic (Figure \ref{topics:family}) might correlate with hospitals restricting visitors during the pandemic.

\begin{figure}[t!]%
    \centering
    {{\includegraphics[width=1\textwidth]{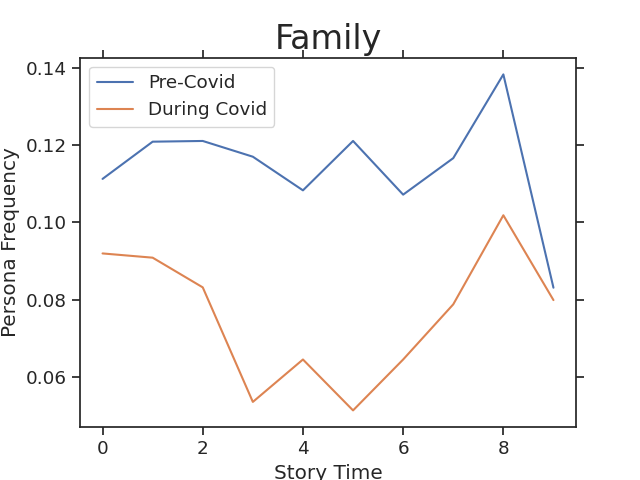}}
    }
    \qquad
    {{\includegraphics[width=1\textwidth]{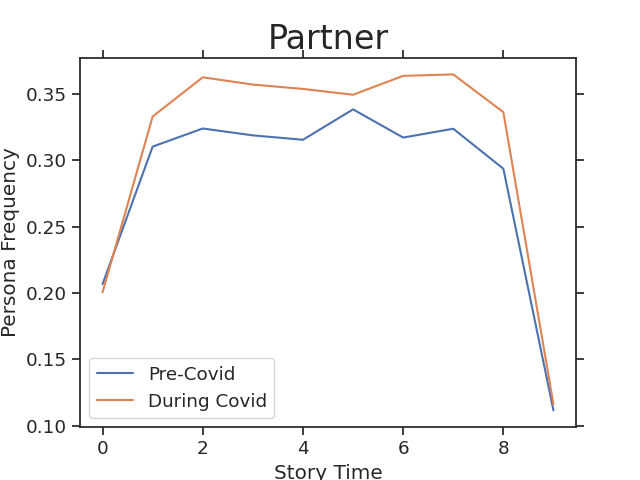}}
    }
    \caption{How often the persona is mentioned on average (y-axis) during the course of a narrative (x-axis).} 
    \label{fig:personas}%
\end{figure}

Our experiments quantifying the personas' presence demonstrate that the healthcare providers were mentioned at similar rates before and during COVID-19, indicating that from the perspective of birthing parents, 
providers' roles in birthing narratives remained consistent.
We notice a significant drop in the family persona's presence (22.8\% , $p < 1^{-7}$) and significant increase in the partner persona's presence (5\%, $p < 0.028$) during the pandemic. Figure \ref{fig:personas} indicates that these differences were usually most apparent from periods 2 to 8 in the stories, which correlates with the time generally spent in the hospital during birthing narratives~\cite{antoniak2019narrative}. This might suggest that partners  supplemented the support missing by families that were unable to visit hospitals before and after delivery. 

\section{Conclusion}
We presented a study demonstrating how parents' self-described birthing experiences significantly changed during COVID-19. Our results indicate that hospital policies may be reflected in birthing narratives. Our work presents a case study in how we can analyze patient experience from their own written narratives and perspectives.

\bibliographystyle{acl}
\bibliography{references}

\begin{thebibliography}{}

\bibitem[\protect\citename{Altman \bgroup et al.\egroup
  }2021]{doi:10.1177/23333936211006397}
Molly~R. Altman, Amelia~R. Gavin, Meghan~K. Eagen-Torkko, Ira
  Kantrowitz-Gordon, Rue~M. Khosa, and Selina~A. Mohammed.
\newblock 2021.
\newblock Where the system failed: The covid-19 pandemic’s impact on
  pregnancy and birth care.
\newblock {\em Global Qualitative Nursing Research}, 8:23333936211006397.
\newblock PMID: 33869668.

\bibitem[\protect\citename{Antoniak \bgroup et al.\egroup
  }2019]{antoniak2019narrative}
Maria Antoniak, David Mimno, and Karen Levy.
\newblock 2019.
\newblock Narrative paths and negotiation of power in birth stories.
\newblock {\em Proceedings of the ACM on Human-Computer Interaction},
  3(CSCW):1--27.

\bibitem[\protect\citename{Baumgartner \bgroup et al.\egroup
  }2020]{Baumgartner2020ThePR}
Jason Baumgartner, Savvas Zannettou, Brian Keegan, Megan Squire, and
  J.~Blackburn.
\newblock 2020.
\newblock The pushshift reddit dataset.
\newblock In {\em ICWSM}.

\bibitem[\protect\citename{Biester \bgroup et al.\egroup
  }2020]{biester-mental-health}
Laura Biester, Katie Matton, Janarthanan Rajendran, Emily~Mower Provost, and
  Rada Mihalcea.
\newblock 2020.
\newblock Quantifying the effects of covid-19 on mental health support forums.
\newblock Computer Science \& Engineering, University of Michigan, sept.

\bibitem[\protect\citename{Blei \bgroup et al.\egroup }2003]{blei2003latent}
David~M Blei, Andrew~Y Ng, and Michael~I Jordan.
\newblock 2003.
\newblock Latent dirichlet allocation.
\newblock {\em the Journal of machine Learning research}, 3:993--1022.

\bibitem[\protect\citename{DeYoung and Mangum}2021]{10.3389/fsoc.2021.611212}
Sarah~E. DeYoung and Michaela Mangum.
\newblock 2021.
\newblock Pregnancy, birthing, and postpartum experiences during covid-19 in
  the united states.
\newblock {\em Frontiers in Sociology}, 6:12.

\bibitem[\protect\citename{Findeklee and Morinello}2020]{findeklee2020clinical}
Sebastian Findeklee and Emanuela Morinello.
\newblock 2020.
\newblock Clinical implications and economic effects of the corona virus
  pandemic on gynaecology, obstetrics and reproductive medicine in
  germany-learning from italy.
\newblock {\em Minerva ginecologica}.

\bibitem[\protect\citename{Goer}2020]{goercovid}
Henci Goer.
\newblock 2020.
\newblock Covid-19: Should you agree to elective induction of labor?

\bibitem[\protect\citename{Gutschow and
  Davis-Floyd}2021]{10.3389/fsoc.2021.655401}
Kim Gutschow and Robbie Davis-Floyd.
\newblock 2021.
\newblock The impacts of covid-19 on us maternity care practices: A followup
  study.
\newblock {\em Frontiers in Sociology}, 6:108.

\bibitem[\protect\citename{Mari{\~n}o-Narvaez \bgroup et al.\egroup
  }2021]{marino2021giving}
Carolina Mari{\~n}o-Narvaez, Jose~A Puertas-Gonzalez, Borja Romero-Gonzalez,
  and Maria~Isabel Peralta-Ramirez.
\newblock 2021.
\newblock Giving birth during the covid-19 pandemic: The impact on birth
  satisfaction and postpartum depression.
\newblock {\em International Journal of Gynecology \& Obstetrics},
  153(1):83--88.

\bibitem[\protect\citename{McCallum}2002]{McCallumMALLET}
Andrew~Kachites McCallum.
\newblock 2002.
\newblock Mallet: A machine learning for language toolkit.
\newblock http://mallet.cs.umass.edu.

\bibitem[\protect\citename{Pilkington and Bedford-Dyer}2021]{mothers-reddit}
Pamela~D. Pilkington and Isabella Bedford-Dyer.
\newblock 2021.
\newblock Mothers' worries during pregnancy: A content analysis of reddit
  posts.
\newblock {\em The Journal of Perinatal Education}, 30(2):98--107.

\bibitem[\protect\citename{Pilkington and Rominov}2017]{fathers-reddit}
Pamela~D. Pilkington and Holly Rominov.
\newblock 2017.
\newblock Fathers’ worries during pregnancy: A qualitative content analysis
  of reddit.
\newblock {\em The Journal of Perinatal Education}, 26(4):208--218.

\bibitem[\protect\citename{R\"{o}der \bgroup et al.\egroup
  }2015]{10.1145/2684822.2685324}
Michael R\"{o}der, Andreas Both, and Alexander Hinneburg.
\newblock 2015.
\newblock Exploring the space of topic coherence measures.
\newblock In {\em Proceedings of the Eighth ACM International Conference on Web
  Search and Data Mining}, WSDM '15, page 399–408, New York, NY, USA.
  Association for Computing Machinery.

\bibitem[\protect\citename{Santosh \bgroup et al.\egroup
  }2020]{santosh-etal-2020-detecting}
Roshan Santosh, H.~Schwartz, Johannes Eichstaedt, Lyle Ungar, and
  Sharath~Chandra Guntuku.
\newblock 2020.
\newblock Detecting emerging symptoms of {COVID}-19 using context-based
  {T}witter embeddings.
\newblock In {\em Proceedings of the 1st Workshop on {NLP} for {COVID}-19 (Part
  2) at {EMNLP} 2020}, Online, December. Association for Computational
  Linguistics.

\bibitem[\protect\citename{Tabak and Purver}2020]{tabak-purver-2020-temporal}
Tom Tabak and Matthew Purver.
\newblock 2020.
\newblock Temporal mental health dynamics on social media.
\newblock In {\em Proceedings of the 1st Workshop on {NLP} for {COVID}-19 (Part
  2) at {EMNLP} 2020}, Online, December. Association for Computational
  Linguistics.

\bibitem[\protect\citename{Taylor and Letham}2018]{taylor2018forecasting}
Sean~J Taylor and Benjamin Letham.
\newblock 2018.
\newblock Forecasting at scale.
\newblock {\em The American Statistician}, 72(1):37--45.

\bibitem[\protect\citename{Thukral \bgroup et al.\egroup
  }2020]{thukral-etal-2020-identifying}
Sachin Thukral, Suyash Sangwan, Arnab Chatterjee, and Lipika Dey.
\newblock 2020.
\newblock Identifying pandemic-related stress factors from social-media posts
  {--} {E}ffects on students and young-adults.
\newblock In {\em Proceedings of the 1st Workshop on {NLP} for {COVID}-19 (Part
  2) at {EMNLP} 2020}, Online, December. Association for Computational
  Linguistics.

\bibitem[\protect\citename{Vazquez-Vazquez \bgroup et al.\egroup
  }2021]{vazquez2021impact}
Adriana Vazquez-Vazquez, Sarah Dib, Emeline Rougeaux, Jonathan~C Wells, and
  MS~Fewtrell.
\newblock 2021.
\newblock The impact of the covid-19 lockdown on the experiences and feeding
  practices of new mothers in the uk: Preliminary data from the covid-19 new
  mum study.
\newblock {\em Appetite}, 156:104985.

\bibitem[\protect\citename{Wolohan}2020]{wolohan-2020-estimating}
JT~Wolohan.
\newblock 2020.
\newblock Estimating the effect of {COVID-19} on mental health: Linguistic
  indicators of depression during a global pandemic.
\newblock In {\em Proceedings of the 1st Workshop on {NLP} for {COVID-19} at
  {ACL} 2020}, Online, July. Association for Computational Linguistics.

\end{thebibliography}

\appendix

\end{document}